\documentclass[11pt]{article}

\usepackage[preprint]{acl}

\usepackage{times}
\usepackage{latexsym}

\usepackage[T1]{fontenc}
\usepackage[utf8]{inputenc}
\usepackage{hyperref}
\usepackage{url}
\usepackage{microtype}
\usepackage{inconsolata}
\usepackage{graphicx}
\usepackage{booktabs} 

\usepackage{amsfonts}
\usepackage{nicefrac}
\usepackage{microtype}
\usepackage{xcolor}
\usepackage{amsmath, amsthm, amssymb}
\usepackage{graphicx}
\usepackage[ruled,vlined,linesnumbered]{algorithm2e}
\usepackage{mathtools}
\usepackage[table]{xcolor}
\usepackage{colortbl}
\usepackage{caption}
\usepackage{wrapfig}
\usepackage{pifont}
\usepackage{colortbl}
\usepackage{lipsum}
\definecolor{mucolor}{RGB}{220, 240, 220}

\newtheorem{theorem}{Theorem}

\newcommand{\algofont}{\fontsize{9.5}{11.5}\selectfont}
\SetAlFnt{\algofont}\SetAlCapFnt{\algofont}\SetAlCapNameFnt{\algofont}

\title{Preserving Utility in Continual LLM Unlearning \\ via Orthogonal Projections}

\author{
  \textbf{Juan Belieni\textsuperscript{1}},
  \textbf{Ana Carolina Erthal\textsuperscript{1}},
  \textbf{Eliezer de Souza da Silva\textsuperscript{2}},
  \textbf{Diego Mesquita\textsuperscript{1}}
\\
  \textsuperscript{1}School of Applied Mathematics, Getulio Vargas Foundation
\\
  \textsuperscript{2}University of Coimbra, CISUC/LASI, DEI
}

\renewcommand{\paragraph}[1]{\textbf{#1}}

\begin{document}
\maketitle

\begin{abstract}
Unlearning concerns the task of removing the influence of a sensitive portion of the training data, such as protected, copyrighted or private text, from a model without retraining it from scratch. While unlearning requests happen continuously during a model's lifetime, the vast majority of LLM unlearning methods are ill-suited for iterated removal, yielding steep decay in forgetting and utility upon sequential unlearning. This decay stems from interference between successive updates, as each new deletion can undo earlier ones or push the model further along directions already degraded by them.
To address this, we introduce Continual Orthogonal Projection (COP), a gradient-projection scheme that restricts each update to the orthogonal complement of activation subspaces cached from prior forget sets. This approach provably guarantees first-order non-interference with earlier deletions. 
COP stores only a low-rank basis per layer, requiring neither past gradients nor previous forget sets, and plugs into unlearning methods that update the model through gradient steps on a differentiable loss without modifying its objective. Experiments on standard LLM unlearning benchmarks (TOFU and RWKU) show that COP preserves utility across sequential requests without weakening forgetting, preventing the collapse of standard objectives and forgetting more effectively than continual-specific baselines.
\end{abstract}
\section{Introduction}

Large language models (LLMs) are trained on web-scale corpora that inevitably include private, copyrighted, or harmful content, which they can memorize and reproduce~\citep{carlini2021extracting,carlini2023quantifying}.
Model providers therefore increasingly face requests to remove specific data from trained models~\citep{shi2025muse}. Because retraining from scratch is prohibitively expensive at this scale, machine unlearning seeks to remove the influence of selected training data directly from a trained model~\citep{cao2015making,bourtoule2021machine,guo2020certified}, erasing targeted information while preserving general capabilities. Long-lived deployments receive these requests sequentially over time, a setting referred to as continual unlearning, in which each new deletion must preserve the effect of earlier ones without access to the data they targeted, while also preventing utility losses from accumulating across successive requests.

Most LLM unlearning methods optimize a gradient-based forget objective, such as gradient ascent (GA)~\citep{jang-etal-2023-knowledge}, negative preference optimization (NPO)~\citep{zhang2024negative}, or its reference-free variant SimNPO~\citep{fan2025simplicity}, alongside a retain regularizer that balances deletion against utility preservation. These methods are designed for isolated requests, and applying them in sequence lets successive updates interfere with one another. Each step can undo earlier deletions or further degrade what they removed, and the retain set alone cannot stop the resulting utility erosion as the request stream grows~\citep{gao2025large}.

Methods designed for the continual setting either move unlearning out of the base model or separate the parameters that each request modifies. O$^3$~\citep{gao2025large} follows the first strategy, freezing the base model and learning all deletions in a shared LoRA adapter, applied at inference according to an out-of-distribution detector. Its behavior is thus contingent on detection errors, and the targeted information remains recoverable from the base weights. ALKN~\citep{wuerkaixi2025adaptive} follows the second, updating the weights through task vectors, encouraging each request to modify a distinct subset of parameters and reducing unlearning intensity where earlier requests already acted. However, separating the parameters each request modifies does not separate their effects. Since every input is processed by the whole network, a new deletion can still interfere with earlier ones by changing the model's behavior on previously forgotten data. In our experiments, neither O$^3$ nor ALKN maintains both forgetting and utility as requests accumulate.

We address this interference by formalizing continual unlearning as a constrained sequential empirical risk minimization problem in which each update must not substantially alter previously satisfied unlearning objectives. Rather than enforcing these constraints through stored parameter gradients, we exploit a layer-wise factorization of the gradient that allows them to be enforced through orthogonality conditions in the much lower-dimensional input activation space, avoiding the need to store past gradients. This results in \textbf{Continual Orthogonal Projection (COP)}, a lightweight plug-in that projects each gradient update onto the orthogonal complement of low-rank subspaces spanned by cached activations from prior forget sets.

Once a request has been processed, we treat its level of forgetting as settled, so later updates have no reason to move along its directions. Further forget steps in these directions would erase more than was accepted at an additional cost in utility, while retain steps could reinstate the forgotten content. COP therefore projects the full update, retain gradient included, so the retain loss repairs utility only in directions orthogonal to earlier deletions. Because it acts only on the gradient, COP works with any unlearning method that updates the model through gradient steps on a differentiable loss, requiring no changes to its objective and no classifier at inference time. On TOFU and RWKU, COP preserves more utility than the base objectives without weakening forgetting, and outperforms continual unlearning baselines.\looseness=-1

In summary, our \textbf{contributions} are:
\begin{enumerate}
    \setlength{\itemsep}{0pt}
    \item a formalization of continual unlearning as a constrained sequential empirical risk minimization (ERM) problem, deriving a first-order interference condition that motivates Continual Orthogonal Projection for any differentiable unlearning objective;
    \item a gradient projection scheme which protects prior deletions by projecting updates away from cached activation subspaces, with first-order non-interference guarantees, no additional asymptotic cost, and memory that scales linearly with each layer's input dimension;
    \item an empirical evaluation showing that COP-augmented methods retain more utility than their base objectives and improve the forget-utility trade-off over continual baselines.
\end{enumerate}

\section{Background}

\paragraph{Continual LLM unlearning.}
Starting from a pretrained language model with parameters $\theta_0$, continual unlearning requires sequentially processing a stream of $T$ deletion requests.
At each step $k$, the model receives a forget set $\mathcal{F}_k$ of prompt-completion pairs $z=(p,c)$ whose information must be removed from the model.
Data from previous forget sets are not accessible during step $k$.
 
We define the cumulative forget set up to step $k$ as $\mathcal{F}_{\leq k} = \bigcup_{i=1}^k \mathcal{F}_i$.
To preserve model utility throughout unlearning, we assume access to a fixed retain set $\mathcal{R} \subset \mathcal{D}$, where $\mathcal{D}$ denotes the original training data, disjoint from all forget sets. The retain set serves as a proxy for the knowledge and behavior that unlearning should leave intact.

Beyond O$^3$ and ALKN, FIT~\citep{xu2026fit} filters redundant requests and adapts the unlearning algorithm and target layers to each one, while CURaTE~\citep{bae-etal-2026-curate} leaves the weights untouched and refuses queries that a sentence embedder judges similar to stored forget requests, so the targeted information remains in the parameters.

\paragraph{Unlearning objectives.}
\label{sec:unlearning-objectives}
At unlearning step $k$, optimization is driven by two losses defined on disjoint subsets. The forget loss $\mathcal{L}^{F}_k$ is computed on the current forget set $\mathcal{F}_k$ and encourages removal of targeted information. The retain loss $\mathcal{L}^{R}_k$ is computed on the retain set $\mathcal{R}$ and regularizes the model to preserve utility. We combine them as $\mathcal{L}_k = \mathcal{L}^{F}_k + \lambda \mathcal{L}^{R}_k$, where $\lambda \geq 0$ controls the forget-retain trade-off.
Other methods instead map the target knowledge to a generic replacement~\citep{eldan2023whos}, perturb its internal representations~\citep{li2024wmdp}, or subtract a task vector learned on the forget data~\citep{ilharco2023editing}.

\paragraph{Layer-wise gradient factorization.}
Consider a linear layer with weights $W \in \mathbb{R}^{d_\mathrm{out}\times d_\mathrm{in}}$, input activation $x \in \mathbb{R}^{d_\mathrm{in}}$, and pre-activation $h = W x$. We omit layer indices, writing $W$ and $G$ for layer weights and gradients, and $\theta$ and $g$ for their full-model counterparts. The per-sample gradient of any differentiable loss $\mathcal{L}$ factorizes as $\nabla_{W}\mathcal{L} \;=\; \delta x^\top$, $\delta \;:=\; \frac{\partial \mathcal{L}}{\partial h} \in \mathbb{R}^{d_\mathrm{out}}$.
Averaging over a forget set $\mathcal{F}_i$, the forget loss gradient with respect to $W$ is
\begin{equation}
  \label{eq:layer_block}
  G_i^F
  \;=\;
  \frac{1}{|\mathcal{F}_i|}\sum_{z\in\mathcal{F}_i}
  \delta_i(z)\,x_i(z)^\top,
\end{equation}
where $x_i(z)$ is the layer input produced by $z \in \mathcal{F}_i$ and $\delta_i(z)$ is the corresponding derivative of the forget loss with respect to $h$.
 
Continual learning methods exploit this structure to protect earlier tasks. OGD~\citep{farajtabar2020orthogonal} projects each new gradient orthogonally to stored gradients of earlier tasks, whereas GPM~\citep{saha2021gradient} stores a low-rank basis of each layer's input activations and projects updates orthogonally to it. COP adopts GPM's activation-space projection but builds the subspaces from past forget sets, so the quantity it protects is the value each earlier unlearning objective reached. Prior unlearning methods use projection either within a single request to protect retained data, as GU~\citep{zhou2025geometricdisentangelment} does for LLMs, or across requests in vision models, as in FG-OrIU~\citep{feng2025fgoriu}; COP targets continual LLM unlearning and leaves the unlearning objective unchanged. Appendix~\ref{app:related} discusses related work in more detail.

\section{Method}

\begin{figure*}
    \centering
    \includegraphics[trim=8em 18em 8em 18em, clip, width=\textwidth]{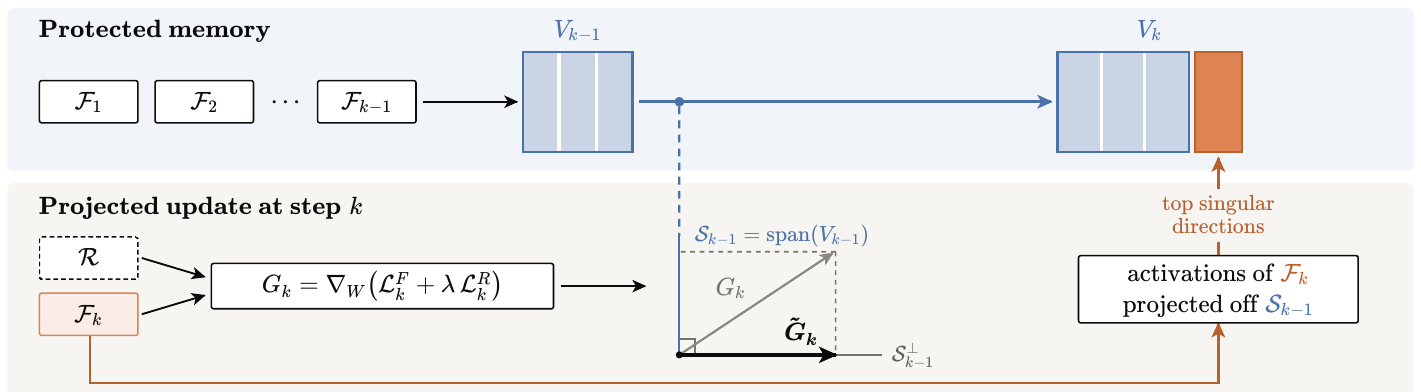}
    \caption{Overview of COP for one linear layer with weights $W$. The input activations of earlier forget sets are summarized by a compact basis $V_{k-1}$, which marks the directions where earlier forgetting lives. At request $k$, the gradient of the unlearning objective on $\mathcal{F}_k$, optionally combined with a retain term on $\mathcal{R}$, is projected away from these directions, so the update cannot change how the layer responds to earlier forget inputs. After training, the dominant new activation directions of $\mathcal{F}_k$ are compressed and added to the basis, giving $V_k$ for the next request.}
    \label{fig:diagram}
\end{figure*}

We cast continual unlearning as a constrained sequential empirical risk minimization problem, seeking parameters that achieve forgetting on $\mathcal{F}_k$, preserve utility on retain data, and avoid interfering with earlier requests, without access to previous forget sets. This problem is formalized and approximated to first order in Section~\ref{sec:erm}, and the resulting constraint is shown to be cheaply enforceable through cached activation subspaces in Section~\ref{sec:activation-projection}. Figure~\ref{fig:diagram} provides an overview of the method.
 
\subsection{Continual unlearning as a constrained ERM problem}
\label{sec:erm}
Each request is optimized on $\mathcal{F}_k$ and $\mathcal{R}$ alone, yet it modifies the same parameters that encode earlier deletions, so earlier forget losses can drift, either restoring removed information or eroding utility. We make this interference explicit by constraining each request to keep earlier forget objectives near their previous values, and derive from the first-order form of these constraints the condition that Section~\ref{sec:activation-projection} enforces from cached activations.

Formally, at step $k$ the forget objective averages a per-example loss over the current forget set,
\begin{equation}
  \label{eq:forget_objective_new}
  \mathcal{L}^{F}_k(\theta)
  \;=\;
  \frac{1}{|\mathcal{F}_k|}\sum_{(p,c)\in \mathcal{F}_k}\ell(\theta; p,c),
\end{equation}
where $\ell$ is any differentiable unlearning loss, with gradient ascent cast as a minimization. The step-$k$ loss is then $\mathcal{L}_k = \mathcal{L}^F_k + \lambda\mathcal{L}^R_k$, as in Section~\ref{sec:unlearning-objectives}.

\paragraph{Preserving earlier requests.}
Let $\theta_i$ denote the parameters after completing step $i$. Later updates should keep each earlier forget objective $\mathcal{L}^F_i$ within a tolerance $\epsilon_i$ of its value at $\theta_i$, in either direction. Increasing $\mathcal{L}^F_i$ partially reverts the deletion, while decreasing it further over-unlearns concepts that were already removed, which can degrade utility. We therefore consider the anchored problem
\begin{equation}
  \label{eq:constrained_erm}
  \min_{\theta}\;\mathcal{L}_k(\theta)
  \quad\text{s.t.}\quad
  \bigl|\mathcal{L}^{F}_i(\theta)-\mathcal{L}^{F}_i(\theta_i)\bigr|\le \epsilon_i\ \ \forall i<k .
\end{equation}
For these anchors to remain well defined, any reference model inside $\mathcal{L}$, such as NPO's reference policy~\citep{zhang2024negative}, must remain the one used during step $i$, since otherwise $\mathcal{L}^F_i$ itself changes across requests.
Since $\mathcal{F}_{\leq k-1}$ is unavailable at step $k$, this problem cannot be solved directly, and we instead use it to derive a tractable update rule.

\paragraph{First-order approximation.}
Let $\theta = \theta_{k-1} + \Delta\theta$ denote a local update to the parameters $\theta_{k-1}$, produced by an optimization step. For each prior step $i < k$, let $a_i := \mathcal{L}^{F}_i(\theta_{k-1})-\mathcal{L}^{F}_i(\theta_i)$ be the drift accumulated before step $k$ and $g_i^F := \nabla_\theta \mathcal{L}^{F}_i(\theta_{k-1})$. A first-order Taylor expansion yields
\begin{equation}
  \label{eq:first_order_change_new}
  \mathcal{L}^{F}_i(\theta_{k-1}+\Delta\theta)-\mathcal{L}^{F}_i(\theta_i)
  \approx a_i + \langle g_i^F, \Delta\theta \rangle,
\end{equation}
so, for sufficiently small steps, the anchored constraint is approximated by the linear conditions
\begin{equation}
  \label{eq:linear_constraints}
  -\epsilon_i - a_i \;\le\; \langle g_i^F,\Delta\theta\rangle \;\le\; \epsilon_i - a_i,
  \quad \forall i<k.
\end{equation}
The homogeneous condition $\langle g_i^F,\Delta\theta\rangle = 0$ leaves the drift unchanged to first order. Imposing this condition therefore keeps the linearized constraint satisfied whenever the current iterate already satisfies $|a_i|\le\epsilon_i$, but it cannot correct already accumulated drift.
Applying a similar Taylor expansion to the current objective $\mathcal{L}_k$ and enforcing the linearized constraints leads to the proximal problem
\begin{equation}
  \label{eq:qp_update}
  \min_{\Delta\theta}\;\; \langle g_k, \Delta\theta \rangle + \tfrac{1}{2\eta}\|\Delta\theta\|^2
  \quad\text{s.t.}\quad \eqref{eq:linear_constraints},
\end{equation}
where $g_k:=\nabla_\theta \mathcal{L}_k(\theta_{k-1})$ is the gradient of the combined objective and the term $\nicefrac{1}{2\eta} \|\Delta\theta\|^2$, with $\eta>0$, penalizes large steps. Without constraints, its solution is the gradient step $\Delta\theta = -\eta g_k$.

\subsection{Continual Orthogonal Projection via cached activations} \label{sec:activation-projection}

Solving Eq.~\eqref{eq:qp_update} directly requires the gradients $g^F_i$ of all previous forget objectives at the current parameters, which cannot be recomputed without $\mathcal{F}_i$ and are impractical to store. We instead show that the linearized constraints can be enforced layer-wise through a cheaper sufficient condition in input activation space. This yields a low-rank projection rule that requires no past gradients, and an incremental basis construction that accumulates protected directions across requests at bounded memory cost.

Since $\langle g^F_i, \Delta\theta \rangle$ is the sum of the per-layer Frobenius inner products $\langle G^F_i, \Delta W \rangle_F$, it suffices to work with one linear layer at a time.
Using Eq.~\eqref{eq:layer_block} together with the identity $\langle u v^\top, M\rangle = u^\top M v$, the contribution of a layer is
\begin{equation}
  \label{eq:trace_identity}
  \langle G^F_i, \Delta W\rangle
  \;=\;
  \frac{1}{|\mathcal{F}_i|}\sum_{z \in \mathcal{F}_i}
  \delta_i(z)^\top\bigl(\Delta W\, x_i(z)\bigr),
\end{equation}
with $x_i(z)$ and $\delta_i(z)$ as in Eq.~\eqref{eq:layer_block}, evaluated at $\theta_{k-1}$. For a language model, each example contributes one such term per token position.
The right-hand side of Eq.~\eqref{eq:trace_identity} depends on $\Delta W$ only through its action on the inputs $x_i(z)$. In particular, letting $X_i = \{x_i(z): z\in\mathcal{F}_i\}$,
\begin{equation}
  \label{eq:activation_condition}
  \Delta W\, x = 0 \quad \forall x \in X_i
  \;\Longrightarrow\;
  \langle G^F_i, \Delta W\rangle = 0.
\end{equation}
The converse does not hold, since a zero inner product only requires the terms of Eq.~\ref{eq:trace_identity} to cancel in aggregate rather than individually.
If Eq. \ref{eq:activation_condition} holds at every trainable layer, summing over them gives $\langle g^F_i,\Delta\theta\rangle=0$, so by \eqref{eq:first_order_change_new}, the drift then remains $a_i$ to first order and the linearized constraint \eqref{eq:linear_constraints} holds whenever the current iterate satisfies $|a_i|\le\epsilon_i$.

The activation-nulling condition $\Delta W x = 0$ for all $x \in X_i$ is therefore sufficient for the linearized constraint and can be enforced without parameter-space gradients. It is also stronger, holding for every example rather than on average and for every differentiable loss rather than for $\mathcal{L}$ alone. Being loss-agnostic and defined only on input activations, it can be encoded by a low-rank basis in input activation space rather than in parameter space.
 
\paragraph{Projection rule.}
Having established activation nulling as a sufficient condition for first-order protection, we can state the projection rule explicitly.

Let $\mathcal{S}_{k-1} \subseteq \mathbb{R}^{d_\mathrm{in}}$ be a subspace containing the protected activations $X_{<k} = \bigcup_{i=1}^{k-1} X_i$, and let $V_{k-1}$ have orthonormal columns spanning $\mathcal{S}_{k-1}$ with corresponding projector $P_{k-1} = V_{k-1}V_{k-1}^\top$. Given the layer gradient $G_k=\nabla_W\mathcal{L}_k$ of the step-$k$ loss, which contains both the forget and the retain term, we form the projected gradient
\begin{equation}
\label{eq:layer_projection}
\boxed{
  \tilde{G}_k
  =
  G_k(I - P_{k-1}),
}
\end{equation}
and update
$W \leftarrow W - \eta \tilde G_k$.
 
For any $x \in \mathcal{S}_{k-1}$, the projected gradient satisfies
$\tilde G_kx=0$. Hence, the corresponding parameter update also acts as zero on protected activations, $\Delta W\, x = -\eta \tilde G_k x = 0$.

Condition~\eqref{eq:activation_condition} therefore holds exactly for every protected activation contained in $\mathcal{S}_{k-1}$, and whenever $X_i \subseteq \mathcal{S}_{k-1}$ at every layer, summing over layers gives $\langle g^F_i,\Delta\theta\rangle=0$. The drift $a_i$ of every such previous request is thus unchanged to first order. With the linearized constraints~\eqref{eq:linear_constraints} replaced by the stronger condition $\Delta W V_{k-1}=0$ at every layer, the solution of Eq.~\eqref{eq:qp_update} is exactly the projected step.

The protection provided by the projection rule holds only to first order and evaluates the protected activations at $\theta_{k-1}$, whereas COP caches $X_i$ once, at $\theta_i$. On fully covered examples, Theorem~\ref{thm:exact} shows that the guarantee is exact, leaving the network's outputs unchanged for any step size.

\begin{theorem}[Exact persistence on protected examples]\label{thm:exact}

Suppose that every trainable parameter is a linear weight, that the subspaces are nested, $\mathcal{S}_i\subseteq\mathcal{S}_{n-1}$ for any step $n > i$, and that every update satisfies $\Delta W\,V_{n-1}=0$ at every layer. 
If every layer input of an example $z$ at $\theta_i$ lies in $\mathcal{S}_i$, then the model output satisfies \[f_{\theta_k}(z)=f_{\theta_i}(z)\] for every $k\ge i$, for any step sizes.
\end{theorem}

Under the hypotheses of Theorem~\ref{thm:exact}, every layer input on a covered example is unchanged after step $i$, so the activations cached at $\theta_i$ coincide with those required by Eq.~\eqref{eq:trace_identity}, and $a_i = 0$ for every fully covered request, for any step size and number of later requests. The hypotheses amount to three requirements. First, the projection constrains the parameter change applied, so with an adaptive optimizer such as Adam it must act on the final step rather than on the raw gradient, since elementwise preconditioning and decoupled weight decay do not preserve $\Delta W V_{k-1}=0$. Second, every trainable parameter is a projected linear weight: embeddings, normalization parameters, and biases are frozen, and a weight shared between operations, such as an output projection tied to the input embedding, is constrained at every use. Third, every layer input of a protected example lies in the protected subspace.

In practice, the third requirement holds only approximately. Compression with $\tau<1$ and the per-request cap $r_{\max}$ discard part of the activation energy of each request, examples outside $\mathcal{F}_i$, such as paraphrases of forgotten content, are not covered, and low-precision arithmetic leaves a small residual in each update. The projection does not control the interference these effects induce, so earlier requests can drift along the unprotected directions, and this drift can accumulate over later requests.

\paragraph{Incremental basis construction.}
Starting from $V_0=\emptyset$, after completing step $k$ we collect the layer input activations $A_k$ on $\mathcal{F}_k$ at $\theta_k$. Before contributing to the basis, these activations are pre-projected against the current basis $V_{k-1}$:
\begin{equation}
  \label{eq:preprojection}
  \widehat A_k
  \;=\;
  \bigl(I - P_{k-1}) \, A_k.
\end{equation}

\begin{algorithm}[t]
\caption{Continual Orthogonal Projection}
\label{alg:COP}
\KwIn{Initial parameters $\theta_0$, stream $\{\mathcal{F}_k\}_{k=1}^T$, retain set $\mathcal{R}$, loss $\mathcal{L}$, threshold $\tau$, rank cap $r_{\max}$, learning rate $\eta$, retain weight $\lambda$, steps per request $m$}
$V_0 \leftarrow \emptyset$, $\theta \leftarrow \theta_0$
 
\For{$k = 1, \dots, T$}{
    $P_{k-1} \leftarrow V_{k-1}V_{k-1}^\top$\;
    \For{$s = 1, \dots, m$}{
        $G_k \leftarrow \nabla_W\bigl(\mathcal{L}^F_k + \lambda\,\mathcal{L}^R_k\bigr)(\theta)$\;
        
        $\tilde{G}_k \leftarrow G_k\,(I - P_{k-1})$\;
        
        $W \leftarrow W - \eta\,\tilde{G}_k$\;
    }
    
    $\theta_k \leftarrow \theta$\;
    
    Collect $A_k$ on $\mathcal{F}_k$ under $\theta_k$\;
    
    $A'_k \leftarrow \mathrm{compress}\bigl((I - P_{k-1})A_k,\, \tau,\, r_{\max}\bigr)$\;
    
    $V_k \leftarrow \mathrm{orthonormalize}\bigl([V_{k-1} \mid A'_k]\bigr)$\;
}
\Return $\theta_T$\;
\end{algorithm}

We extract the smallest number of leading left singular vectors of $\widehat A_k$ that retain a fraction $\tau\in(0,1]$ of its squared Frobenius norm, keeping at most $r_{\max}$ of them, and create $V_{k}$ by concatenating these vectors with $V_{k-1}$ followed by a QR re-orthonormalization.
The resulting subspaces are nested, $\mathcal{S}_{k-1}\subseteq\mathcal{S}_k$, as required by Theorem~\ref{thm:exact}.

Algorithm~\ref{alg:COP} summarizes the complete procedure, which alternates between projected training on each request and an update of the protected basis from that request's activations. The activations $A_k$, bases $V_k$, projectors $P_k$, and updates of $W$ are computed separately for each linear layer.

\subsection{Complexity and capacity}
\label{sec:memory}
 
Protecting earlier requests has two costs: the memory to store the protected subspaces, and the update directions removed from later requests. Both are governed by the rank $r$ of the basis $V_k$ at each layer.
 
\paragraph{Memory and compute.}
For each linear layer, COP stores only the basis $V_k$, which has $r\,d_\mathrm{in}$ entries. Neither previous forget sets nor their activations are kept, so memory grows with the rank of the protected subspace rather than with the number or size of the requests. Storing even a single parameter-space gradient would instead require $d_\mathrm{in}\,d_\mathrm{out}$ entries per layer. The projection is computed as $G_k - (G_kV_{k-1})V_{k-1}^\top$, without forming $P_{k-1}$, and adds two light matrix products to each update, which is small compared with a forward and backward pass whenever $r$ is below the number of tokens in a batch. Building the basis requires one additional forward pass over $\mathcal{F}_k$ per request, followed by an SVD and a QR decomposition.
 
\paragraph{Capacity.}
The protected subspace also limits what later requests can change. An update satisfying $\Delta W V_{k-1}=0$ acts only on the $d_\mathrm{in}-r$ input directions outside $\mathcal{S}_{k-1}$, and since the bases are nested, $r$ never decreases from one request to the next. Therefore, each request leaves less room for those that follow, and once $\mathcal{S}_{k-1}$ spans the whole input space of a layer, that layer can no longer be updated. Free directions are also not sufficient on their own: if the activations of the current forget set already lie in $\mathcal{S}_{k-1}$, the projected gradient vanishes at that layer even when $r<d_\mathrm{in}$.
 
The threshold $\tau$ and cap $r_{\max}$ keep this growth in check. Each request adds at most $r_{\max}$ directions per layer, so after $k$ requests $r\le\min(d_\mathrm{in},\,k\,r_{\max})$, and a lower $\tau$ adds fewer directions still. Both leave more of the input space free for later requests, at the price of exact protection: directions discarded by the compression are no longer shielded, and earlier requests may drift along them. Basis compression is thus a way to control capacity and save memory.

\section{Experiments}
\label{sec:experiments}

We evaluate COP on two benchmarks that probe complementary aspects of unlearning in LLMs. TOFU~\citep{maini2024tofu} asks the model to forget fictitious authors whose knowledge was introduced only through fine-tuning, isolating the effect of unlearning on entity-specific facts. RWKU~\citep{jin2024rwku} instead asks it to forget real people whose knowledge was acquired during pretraining, where no controlled fine-tuning stage exists. In both, the forget data is split into $T$ requests $\mathcal{F}_1,\ldots,\mathcal{F}_T$ that every method processes in the same order on a single model, alongside a fixed retain set $\mathcal{R}$, and a successful method must forget each request while preserving general capabilities.

\paragraph{Metrics.}
We measure forgetting with forget ROUGE (FR~$\downarrow$), forget probability (FP~$\downarrow$), exact memorization (EM~$\downarrow$)~\citep{tirumala2022memorization}, and extraction strength (ES~$\downarrow$)~\citep{carlini2021extracting}, reporting FR, ES, and EM on TOFU and FR, FP, and EM on RWKU. After request $k$, they are computed over all requests seen so far, so they also capture whether earlier deletions remain forgotten. Utility is measured by model utility (MU~$\uparrow$), a harmonic mean of benchmark-specific utility measures. Definitions are given in Appendix~\ref{app:metrics}.

\paragraph{Baselines.}
We apply COP to four widely used unlearning objectives. Gradient ascent (GA)~\citep{jang-etal-2023-knowledge} maximizes the likelihood loss on the forget set, NPO~\citep{zhang2024negative} recasts forgetting as preference optimization against a fixed reference model, and SimNPO~\citep{fan2025simplicity} removes the reference model in favor of length-normalized rewards. IDK~\citep{maini2024tofu} replaces the suppression of original answers with fine-tuning toward refusals. Each objective is paired with its COP-augmented counterpart with $r_{\max}=256$. We further compare against ALKN~\citep{wuerkaixi2025adaptive} and O$^3$~\citep{gao2025large}, methods built for continual unlearning. Each method and variant is tuned for the best forget-utility trade-off (Appendix~\ref{app:impl}).

\begin{figure}[t]
    \centering
    \includegraphics[width=\columnwidth]{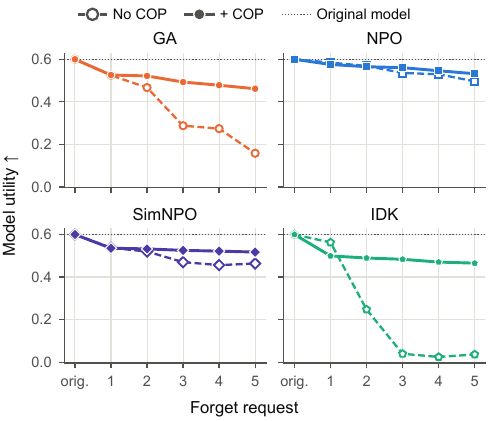}
    \caption{\textbf{COP preserves utility as forget requests accumulate.} Model utility on TOFU (\texttt{forget05}) with Llama-3.2-1B-Instruct after each of the five forget requests, for each base objective (dashed, hollow markers) and its COP-augmented variant (solid, filled markers). The dotted line marks the utility of the original model before unlearning. Utility under COP stays close to its early level as requests accumulate, whereas base GA and IDK collapse and base NPO and SimNPO fall increasingly behind their COP variants in later requests.}
    \label{fig:tofu-utility}
    \vspace{-1em}
\end{figure}

\subsection{TOFU}
\label{sec:tofu}

\paragraph{Setup.}
TOFU~\citep{maini2024tofu} contains question-answer pairs about 200 fictitious authors. We start from the Llama-3.2-1B-Instruct and Llama-3.2-3B-Instruct checkpoints fine-tuned on the full dataset by \citet{dorna2025openunlearning} and split the \texttt{forget05} subset into $T=5$ requests of two authors and 40 question-answer pairs each, using \texttt{retain95} as $\mathcal{R}$. MU is the harmonic mean of answer probability, ROUGE, and truth ratio on the retain set, real authors, and world facts~\citep{maini2024tofu}. Further details are in Appendix~\ref{app:tofu}.

\begin{table*}[t]
\begin{center}
\tiny
\renewcommand{\arraystretch}{1.25}
\setlength{\tabcolsep}{4pt}
\begin{tabular}{l|ccc>{\columncolor{mucolor}}c|ccc>{\columncolor{mucolor}}c|ccc>{\columncolor{mucolor}}c|ccc>{\columncolor{mucolor}}c|ccc>{\columncolor{mucolor}}c}
\hline
\textbf{Method} & \multicolumn{4}{c|}{\bf Task 1} & \multicolumn{4}{c|}{\bf Task 2} & \multicolumn{4}{c|}{\bf Task 3} & \multicolumn{4}{c|}{\bf Task 4} & \multicolumn{4}{c}{\bf Task 5} \\
 & FR$\downarrow$ & ES$\downarrow$ & EM$\downarrow$ & MU$\uparrow$ & FR$\downarrow$ & ES$\downarrow$ & EM$\downarrow$ & MU$\uparrow$ & FR$\downarrow$ & ES$\downarrow$ & EM$\downarrow$ & MU$\uparrow$ & FR$\downarrow$ & ES$\downarrow$ & EM$\downarrow$ & MU$\uparrow$ & FR$\downarrow$ & ES$\downarrow$ & EM$\downarrow$ & MU$\uparrow$ \\
\hline
 Original & 0.92 & 0.81 & 0.98 & 0.599 & 0.91 & 0.83 & 0.98 & 0.599 & 0.81 & 0.73 & 0.97 & 0.599 & 0.82 & 0.72 & 0.97 & 0.599 & 0.83 & 0.73 & 0.97 & 0.599 \\
\hline
 GA & 0.47 & 0.10 & 0.71 & 0.524 & 0.40 & 0.09 & 0.66 & 0.467 & 0.30 & 0.06 & 0.52 & 0.288 & 0.25 & 0.05 & 0.46 & 0.274 & 0.17 & 0.03 & 0.30 & 0.159 \\
 GA+COP & 0.47 & 0.10 & 0.71 & \textbf{0.526} & 0.43 & 0.10 & 0.69 & \textbf{0.521} & 0.35 & 0.08 & 0.65 & \textbf{0.493} & 0.35 & 0.08 & 0.63 & \textbf{0.478} & 0.34 & 0.07 & 0.63 & \textbf{0.461} \\
\arrayrulecolor{gray!50}\hline\arrayrulecolor{black}
 NPO & 0.44 & 0.09 & 0.67 & \textbf{0.584} & 0.38 & 0.09 & 0.67 & \textbf{0.568} & 0.35 & 0.09 & 0.65 & 0.535 & 0.34 & 0.08 & 0.64 & 0.529 & 0.32 & 0.08 & 0.63 & 0.496 \\
 NPO+COP & 0.40 & 0.06 & 0.57 & 0.575 & 0.36 & 0.06 & 0.57 & 0.565 & 0.32 & 0.05 & 0.55 & \textbf{0.560} & 0.31 & 0.05 & 0.55 & \textbf{0.546} & 0.30 & 0.05 & 0.56 & \textbf{0.532} \\
\arrayrulecolor{gray!50}\hline\arrayrulecolor{black}
 SimNPO & 0.43 & 0.11 & 0.72 & \textbf{0.537} & 0.43 & 0.10 & 0.71 & 0.520 & 0.34 & 0.09 & 0.67 & 0.470 & 0.36 & 0.08 & 0.66 & 0.456 & 0.36 & 0.07 & 0.65 & 0.463 \\
 SimNPO+COP & 0.44 & 0.11 & 0.71 & 0.535 & 0.42 & 0.10 & 0.72 & \textbf{0.532} & 0.36 & 0.09 & 0.69 & \textbf{0.525} & 0.38 & 0.09 & 0.69 & \textbf{0.521} & 0.37 & 0.08 & 0.69 & \textbf{0.517} \\
\arrayrulecolor{gray!50}\hline\arrayrulecolor{black}
 IDK & 0.08 & 0.12 & 0.67 & \textbf{0.562} & 0.02 & 0.09 & 0.60 & 0.248 & 0.01 & 0.07 & 0.55 & 0.040 & 0.03 & 0.07 & 0.53 & 0.025 & 0.02 & 0.05 & 0.49 & 0.037 \\
 IDK+COP & 0.29 & 0.14 & 0.73 & 0.499 & 0.18 & 0.10 & 0.65 & \textbf{0.489} & 0.12 & 0.09 & 0.59 & \textbf{0.483} & 0.13 & 0.08 & 0.59 & \textbf{0.470} & 0.10 & 0.07 & 0.59 & \textbf{0.465} \\
\arrayrulecolor{gray!50}\hline\arrayrulecolor{black}
 O$^3$ & 0.59 & 0.56 & 0.89 & 0.598 & 0.54 & 0.61 & 0.92 & 0.598 & 0.50 & 0.51 & 0.90 & 0.597 & 0.51 & 0.50 & 0.90 & 0.595 & 0.30 & 0.36 & 0.84 & 0.576 \\
\arrayrulecolor{gray!50}\hline\arrayrulecolor{black}
 ALKN & 0.43 & 0.07 & 0.60 & 0.534 & 0.34 & 0.05 & 0.45 & 0.338 & 0.09 & 0.04 & 0.21 & 0.018 & 0.05 & 0.03 & 0.10 & 0.007 & 0.02 & 0.03 & 0.06 & 0.004 \\
\hline
\end{tabular}
\end{center}
\caption{\textbf{TOFU results on Llama-3.2-1B-Instruct} (\texttt{forget05}). After each task, forget metrics are computed over all requests seen so far, and model utility (MU) is measured on the retain set and general utility datasets. Bold marks the higher MU within each pair of a base objective and its COP-augmented variant. From the third request onward, every COP variant retains higher utility than its base objective. COP prevents the utility collapse of GA and IDK, whose base versions reach lower EM only as part of that collapse, while forget metrics for NPO and SimNPO remain comparable with and without COP. O$^3$ attains the highest utility but forgets weakly, with EM close to that of the original model. ALKN's forget metrics fall together with its utility, which collapses as requests accumulate.}
\label{tab:tofu-1b}
\end{table*}

\begin{table*}[t]
\begin{center}
\tiny
\renewcommand{\arraystretch}{1.25}
\setlength{\tabcolsep}{4pt}
\begin{tabular}{l|ccc>{\columncolor{mucolor}}c|ccc>{\columncolor{mucolor}}c|ccc>{\columncolor{mucolor}}c|ccc>{\columncolor{mucolor}}c|ccc>{\columncolor{mucolor}}c}
\hline
\textbf{Method} & \multicolumn{4}{c|}{\bf Task 1} & \multicolumn{4}{c|}{\bf Task 2} & \multicolumn{4}{c|}{\bf Task 3} & \multicolumn{4}{c|}{\bf Task 4} & \multicolumn{4}{c}{\bf Task 5} \\
 & FR$\downarrow$ & ES$\downarrow$ & EM$\downarrow$ & MU$\uparrow$ & FR$\downarrow$ & ES$\downarrow$ & EM$\downarrow$ & MU$\uparrow$ & FR$\downarrow$ & ES$\downarrow$ & EM$\downarrow$ & MU$\uparrow$ & FR$\downarrow$ & ES$\downarrow$ & EM$\downarrow$ & MU$\uparrow$ & FR$\downarrow$ & ES$\downarrow$ & EM$\downarrow$ & MU$\uparrow$ \\
\hline
 Original & 0.97 & 0.93 & 0.99 & 0.667 & 0.96 & 0.93 & 0.99 & 0.667 & 0.90 & 0.87 & 0.99 & 0.667 & 0.92 & 0.88 & 0.99 & 0.667 & 0.93 & 0.89 & 0.99 & 0.667 \\
\hline
 GA & 0.55 & 0.23 & 0.80 & \textbf{0.641} & 0.34 & 0.11 & 0.66 & 0.517 & 0.19 & 0.05 & 0.35 & 0.380 & 0.14 & 0.04 & 0.23 & 0.393 & 0.07 & 0.03 & 0.12 & 0.224 \\
 GA+COP & 0.53 & 0.24 & 0.80 & \textbf{0.641} & 0.52 & 0.20 & 0.80 & \textbf{0.634} & 0.43 & 0.18 & 0.77 & \textbf{0.616} & 0.41 & 0.15 & 0.74 & \textbf{0.594} & 0.38 & 0.12 & 0.71 & \textbf{0.566} \\
\arrayrulecolor{gray!50}\hline\arrayrulecolor{black}
 NPO & 0.36 & 0.07 & 0.60 & \textbf{0.658} & 0.30 & 0.07 & 0.62 & \textbf{0.640} & 0.24 & 0.07 & 0.57 & 0.592 & 0.23 & 0.06 & 0.54 & 0.548 & 0.21 & 0.05 & 0.52 & 0.456 \\
 NPO+COP & 0.37 & 0.07 & 0.59 & 0.614 & 0.37 & 0.07 & 0.57 & 0.603 & 0.32 & 0.06 & 0.55 & \textbf{0.593} & 0.33 & 0.05 & 0.55 & \textbf{0.583} & 0.33 & 0.06 & 0.56 & \textbf{0.571} \\
\arrayrulecolor{gray!50}\hline\arrayrulecolor{black}
 SimNPO & 0.52 & 0.12 & 0.75 & 0.613 & 0.41 & 0.11 & 0.72 & 0.611 & 0.36 & 0.09 & 0.69 & 0.569 & 0.37 & 0.08 & 0.68 & 0.558 & 0.36 & 0.08 & 0.66 & 0.554 \\
 SimNPO+COP & 0.49 & 0.12 & 0.76 & \textbf{0.614} & 0.43 & 0.10 & 0.73 & \textbf{0.617} & 0.37 & 0.10 & 0.71 & \textbf{0.608} & 0.37 & 0.09 & 0.71 & \textbf{0.603} & 0.38 & 0.09 & 0.71 & \textbf{0.600} \\
\arrayrulecolor{gray!50}\hline\arrayrulecolor{black}
 IDK & 0.06 & 0.24 & 0.76 & 0.634 & 0.03 & 0.16 & 0.69 & 0.409 & 0.01 & 0.12 & 0.64 & 0.026 & 0.02 & 0.10 & 0.60 & 0.020 & 0.03 & 0.07 & 0.55 & 0.056 \\
 IDK+COP & 0.07 & 0.24 & 0.76 & \textbf{0.635} & 0.05 & 0.17 & 0.71 & \textbf{0.600} & 0.03 & 0.15 & 0.68 & \textbf{0.562} & 0.03 & 0.12 & 0.67 & \textbf{0.521} & 0.04 & 0.11 & 0.67 & \textbf{0.477} \\
\arrayrulecolor{gray!50}\hline\arrayrulecolor{black}
 O$^3$ & 0.63 & 0.68 & 0.93 & 0.664 & 0.58 & 0.72 & 0.94 & 0.664 & 0.55 & 0.65 & 0.94 & 0.663 & 0.57 & 0.70 & 0.95 & 0.663 & 0.34 & 0.51 & 0.91 & 0.639 \\
\arrayrulecolor{gray!50}\hline\arrayrulecolor{black}
 ALKN & 0.34 & 0.05 & 0.42 & 0.436 & 0.09 & 0.04 & 0.16 & 0.084 & 0.04 & 0.03 & 0.14 & 0.042 & 0.04 & 0.03 & 0.16 & 0.040 & 0.04 & 0.03 & 0.17 & 0.040 \\
\hline
\end{tabular}
\end{center}
\caption{\textbf{TOFU results on Llama-3.2-3B-Instruct} (\texttt{forget05}), with the same metrics and notation as Table~\ref{tab:tofu-1b}. COP improves utility at every request for SimNPO and IDK, from the second request onward for GA, and from the third onward for NPO. Base GA and IDK drive forget ROUGE close to zero only as their utility collapses, while their COP variants remain stable at the cost of higher EM. O$^3$ attains the highest utility but fails to forget effectively, with EM above 0.9 at every request. ALKN's utility falls sharply as requests accumulate.}
\label{tab:tofu-3b}
\end{table*}

\paragraph{Results.}
Across both model sizes, Tables~\ref{tab:tofu-1b} and~\ref{tab:tofu-3b} show that COP consistently slows the utility loss that accumulates over sequential requests, while keeping forgetting close to that of the base objectives, and from the third request onwards, every COP-augmented method retains higher utility than its base objective. The gap is largest for GA and IDK, whose base versions collapse, while their COP variants remain comparable to other methods. Their low forget ROUGE follows this collapse, and thus reflects a degraded model rather than effective unlearning. For NPO and SimNPO, whose base versions degrade more gradually, COP still improves final utility. Forgetting remains comparable for SimNPO and improves for NPO on the 1B model, while on the 3B model NPO+COP shows higher FR in exchange for its larger utility gain.

Figure~\ref{fig:tofu-utility} traces this utility gap over the stream for the 1B model. With no earlier deletions to protect at the first request, COP leaves that update unconstrained, and each base objective and its COP variant reach similar utility after it. Trajectories diverge from the second request onwards, as base objectives lose utility with each request while their COP variants decline slowly. This matches the behavior COP is designed for, which is not to reduce the cost of an individual deletion, but to prevent that cost from compounding as requests accumulate.

Finally, among continual baselines, O$^3$ maintains nearly constant utility, but at the cost of not forgetting effectively when compared to the other methods. Both effects likely stem from its out-of-distribution detector, which scales the unlearning adapter by how similar each query is to the forget data. Queries judged dissimilar are answered by the unmodified base model, which preserves utility on retain data, but any forget query that the detector fails to flag is also answered by the base model, and thus remains unforgotten. ALKN shows the opposite behavior, as its utility collapses within the first few requests on both models, so its near-zero forget ROUGE, as for base GA and IDK, reflects a degraded model rather than effective unlearning. Appendix~\ref{app:rank} further shows that, within the ranks we test, increasing the rank cap $r_{\max}$ improves utility without preventing later deletions, since extraction strength stays nearly constant.

\begin{figure}[t]
    \centering
    \includegraphics[width=\columnwidth]{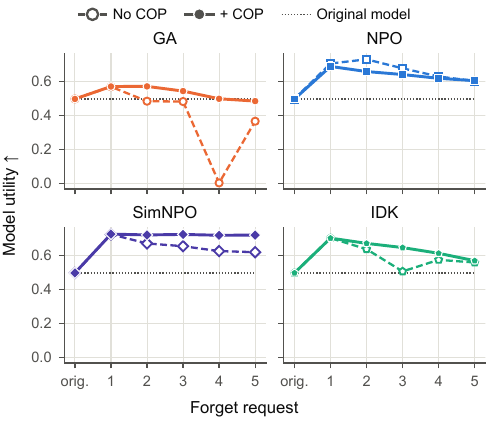}
    \caption{\textbf{COP maintains higher utility on real-world knowledge unlearning.} Model utility on RWKU with Llama-3.2-3B-Instruct after each forget request, for each base objective (dashed, hollow markers) and its COP-augmented variant (solid, filled markers). Utility under COP remains stable as requests accumulate, while base objectives fluctuate or degrade.}
    \label{fig:rwku-utility}
    \vspace{-1em}
\end{figure}

\begin{table*}[t]
\begin{center}
\tiny
\renewcommand{\arraystretch}{1.25}
\setlength{\tabcolsep}{4pt}
\begin{tabular}{l|ccc>{\columncolor{mucolor}}c|ccc>{\columncolor{mucolor}}c|ccc>{\columncolor{mucolor}}c|ccc>{\columncolor{mucolor}}c|ccc>{\columncolor{mucolor}}c}
\hline
\textbf{Method} & \multicolumn{4}{c|}{\bf Task 1} & \multicolumn{4}{c|}{\bf Task 2} & \multicolumn{4}{c|}{\bf Task 3} & \multicolumn{4}{c|}{\bf Task 4} & \multicolumn{4}{c}{\bf Task 5} \\
 & FR$\downarrow$ & FP$\downarrow$ & EM$\downarrow$ & MU$\uparrow$ & FR$\downarrow$ & FP$\downarrow$ & EM$\downarrow$ & MU$\uparrow$ & FR$\downarrow$ & FP$\downarrow$ & EM$\downarrow$ & MU$\uparrow$ & FR$\downarrow$ & FP$\downarrow$ & EM$\downarrow$ & MU$\uparrow$ & FR$\downarrow$ & FP$\downarrow$ & EM$\downarrow$ & MU$\uparrow$ \\
\hline
 Original & 0.78 & 0.34 & 0.70 & 0.497 & 0.73 & 0.31 & 0.67 & 0.497 & 0.74 & 0.30 & 0.66 & 0.497 & 0.75 & 0.32 & 0.68 & 0.497 & 0.72 & 0.32 & 0.66 & 0.497 \\
\hline
 GA & 0.34 & 0.00 & 0.36 & 0.569 & 0.20 & 0.02 & 0.26 & 0.484 & 0.08 & 0.01 & 0.21 & 0.482 & 0.03 & 0.00 & 0.11 & 0.004 & 0.02 & 0.00 & 0.09 & 0.366 \\
 GA+COP & 0.33 & 0.00 & 0.36 & \textbf{0.571} & 0.24 & 0.02 & 0.28 & \textbf{0.571} & 0.18 & 0.01 & 0.26 & \textbf{0.544} & 0.19 & 0.02 & 0.25 & \textbf{0.498} & 0.15 & 0.01 & 0.21 & \textbf{0.485} \\
\arrayrulecolor{gray!50}\hline\arrayrulecolor{black}
 NPO & 0.28 & 0.03 & 0.39 & \textbf{0.706} & 0.24 & 0.03 & 0.37 & \textbf{0.730} & 0.19 & 0.02 & 0.34 & \textbf{0.677} & 0.16 & 0.01 & 0.32 & \textbf{0.630} & 0.12 & 0.01 & 0.30 & 0.603 \\
 NPO+COP & 0.04 & 0.01 & 0.32 & 0.689 & 0.07 & 0.00 & 0.27 & 0.659 & 0.05 & 0.00 & 0.23 & 0.641 & 0.05 & 0.00 & 0.23 & 0.619 & 0.04 & 0.00 & 0.20 & \textbf{0.605} \\
\arrayrulecolor{gray!50}\hline\arrayrulecolor{black}
 SimNPO & 0.27 & 0.29 & 0.58 & 0.724 & 0.22 & 0.20 & 0.51 & 0.671 & 0.19 & 0.15 & 0.45 & 0.655 & 0.16 & 0.11 & 0.44 & 0.626 & 0.13 & 0.11 & 0.42 & 0.620 \\
 SimNPO+COP & 0.26 & 0.29 & 0.58 & \textbf{0.726} & 0.21 & 0.19 & 0.51 & \textbf{0.721} & 0.18 & 0.14 & 0.48 & \textbf{0.724} & 0.16 & 0.11 & 0.48 & \textbf{0.719} & 0.15 & 0.10 & 0.44 & \textbf{0.720} \\
\arrayrulecolor{gray!50}\hline\arrayrulecolor{black}
 IDK & 0.13 & 0.16 & 0.44 & \textbf{0.702} & 0.12 & 0.11 & 0.39 & 0.637 & 0.06 & 0.07 & 0.37 & 0.506 & 0.04 & 0.05 & 0.31 & 0.574 & 0.03 & 0.04 & 0.28 & 0.559 \\
 IDK+COP & 0.13 & 0.16 & 0.44 & 0.702 & 0.09 & 0.10 & 0.36 & \textbf{0.671} & 0.07 & 0.07 & 0.33 & \textbf{0.647} & 0.04 & 0.05 & 0.32 & \textbf{0.613} & 0.04 & 0.04 & 0.29 & \textbf{0.569} \\
\arrayrulecolor{gray!50}\hline\arrayrulecolor{black}
 O$^3$ & 0.12 & 0.06 & 0.49 & 0.273 & 0.09 & 0.06 & 0.46 & 0.197 & 0.06 & 0.04 & 0.45 & 0.083 & 0.04 & 0.03 & 0.50 & 0.081 & 0.03 & 0.03 & 0.45 & 0.062 \\
\arrayrulecolor{gray!50}\hline\arrayrulecolor{black}
 ALKN & 0.50 & 0.02 & 0.41 & 0.070 & 0.46 & 0.02 & 0.37 & 0.089 & 0.36 & 0.04 & 0.37 & 0.134 & 0.35 & 0.05 & 0.38 & 0.162 & 0.32 & 0.05 & 0.37 & 0.156 \\
\hline
\end{tabular}
\end{center}
\caption{\textbf{RWKU results on Llama-3.2-3B-Instruct}. After each task, forget metrics are computed over all requests seen so far, and model utility (MU) is measured on the retained subjects and world facts. Bold marks the higher MU within each pair of a base objective and its COP-augmented variant. COP keeps utility stable across requests, while O$^3$ forgets at the cost of most of its utility and ALKN retains much of the targeted knowledge.}
\label{tab:rwku}
\end{table*}

\subsection{RWKU}
 
\paragraph{Setup.}
RWKU~\citep{jin2024rwku} targets knowledge of 200 real, widely known people that a model acquires during pretraining. We use its question-answering probes (level 2), short questions about a person paired with an answer of a few words. We draw 20 people at random with a fixed seed and group them into $T=5$ requests of four people each, so that $\mathcal{F}_k$ contains all probes of its four people and $\mathcal{R}$ contains the probes of the 180 people who are never forgotten. We use Llama-3.2-3B-Instruct~\citep{grattafiori2024llama} without further fine-tuning, since the targeted knowledge is already present in the pretrained model. MU is the harmonic mean of answer probability and ROUGE on $\mathcal{R}$ and of normalized answer probability and ROUGE on the TOFU world-facts questions~\citep{maini2024tofu}. Further details are in Appendix~\ref{app:rwku}.

\paragraph{Results.}
COP remains effective when the targeted knowledge comes from pretraining, stabilizing utility across the stream while base objectives fluctuate or degrade. Table~\ref{tab:rwku} and Figure~\ref{fig:rwku-utility} show the effect is clearest for GA, whose base version collapses at the fourth request while GA+COP declines more smoothly, and SimNPO and IDK with COP keep utility nearly constant while their base versions decline. NPO shows a different trade-off, as NPO+COP forgets considerably more than NPO while reaching the same final utility.

The continual baselines behave differently than on TOFU, as O$^3$ now forgets effectively but loses most of its utility, while ALKN retains much of the targeted knowledge at low utility, so neither matches the trade-off of COP variants. Note that utility under most runs and every COP variant rises slightly above the original model. This is expected, since the retain loss is computed on $\mathcal{R}$ and MU is partly evaluated on the same distribution.

\section{Conclusion}

We proposed Continual Orthogonal Projection, a continual unlearning method that projects each update away from activation subspaces cached from earlier forget sets.
This leaves earlier unlearning objectives unchanged to first order without storing past data and without modifying the underlying unlearning loss.
Across TOFU and RWKU, COP substantially reduced the utility loss that accumulates over sequential requests while keeping forgetting comparable, offering a practical option for deployed models facing an ongoing stream of deletion requests, without adding inference cost.

\section*{Limitations}

COP leaves earlier forget losses exactly unchanged only on inputs whose activations lie in the cached subspaces, and only if the realized update is projected and every trainable parameter is a linear weight. Our runs relax these conditions, since compression keeps at most $r_{\max}$ directions per request, AdamW and weight decay act after the projection, and embeddings and normalization gains are not projected, so the protection we observe is approximate. The protected subspace also grows with every request, which limits the length of the stream a fixed rank can support. Our evaluation uses a single seed, five requests, models of up to 3B parameters, and forget metrics on the training probes, and does not include relearning or extraction attacks. COP should therefore not be treated as a certified deletion mechanism or as sufficient on its own for legal or regulatory compliance. Finally, unlearning methods are dual-use, since the same procedure could remove safety-relevant knowledge or suppress legitimate information.

\section*{Acknowledgments}

We acknowledge the support of the Fundação Carlos Chagas Filho de Amparo à Pesquisa do Estado do Rio de Janeiro (FAPERJ) (SEI-260003/020348/2025, SEI-260003/020694/2025) and the Conselho Nacional de Desenvolvimento Científico e Tecnológico (CNPq) (404336/2023-0, 305692/2025-9, 445170/2024-7).

E. da Silva's work is funded by Portuguese national funds through FCT – Foundation for Science and Technology, I.P., within the scope of the research unit UID/00326 - Centre for Informatics and Systems of the University of Coimbra.

We used Claude for drafting and editing parts of the text and for literature search, and Claude Code for coding. The authors reviewed and revised all outputs and take full responsibility for the content.

\bibliography{custom,anthology-1,anthology-2}

\appendix

\section{Related Work}
\label{app:related}
 
\paragraph{LLM unlearning.}
Exact unlearning retrains the model without the forget data~\citep{cao2015making, bourtoule2021machine}. Certified variants bound how far the unlearned model can be from a retrained one~\citep{guo2020certified}.
Both become impractical at LLM scale, so most work fine-tunes the deployed model directly.
Gradient ascent~\citep{jang-etal-2023-knowledge} is the simplest approach, as it maximizes the next-token loss on the forget set, but its loss is unbounded and quickly erodes general utility. 
NPO~\citep{zhang2024negative} adapts DPO~\citep{rafailov2023direct} to the case where only rejected responses are available, penalizing the likelihood ratio between the current and the reference model on forget answers. SimNPO~\citep{fan2025simplicity} removes the need for a reference model and uses the length-normalized log-likelihood of the forget answer instead, and IDK~\citep{maini2024tofu} instead trains the model to answer forget queries with a refusal such as ``I don't know''. Its preference-based variant applies DPO with the refusal as the preferred response and the original answer as the rejected one.
 
Other methods map the target knowledge to a generic replacement~\citep{eldan2023whos}, perturb internal representations of it~\citep{li2024wmdp}, or subtract a task vector learned on the forget data~\citep{ilharco2023editing}.
 
\paragraph{Continual unlearning.}
When deletion requests arrive one after another, each update can damage general utility and also undo earlier deletions, and these effects compound over the sequence.  O$^3$~\citep{gao2025large} trains a shared LoRA adapter under an orthogonality regularizer and uses an out-of-distribution detector to decide at inference time how strongly the adapter is applied. ALKN~\citep{wuerkaixi2025adaptive} works on the weights directly, combining task-vector negation with gradient masks that discourage requests from reusing the same parameters. FIT~\citep{xu2026fit} filters redundant requests and, for each remaining one, selects an unlearning algorithm according to its estimated importance and restricts the update to attributed layers. CURaTE~\citep{bae-etal-2026-curate} leaves the weights untouched and refuses queries that a sentence embedder judges similar to stored forget requests, so the targeted information remains in the parameters.

\paragraph{Gradient projection in continual learning.}
Continual learning faces the same interference between sequential updates, with earlier tasks to retain instead of earlier deletions to preserve. OGD~\citep{farajtabar2020orthogonal} projects each new gradient orthogonally to gradients stored from earlier tasks. GPM~\citep{saha2021gradient} stores, for each layer, a low-rank basis of the input activations of earlier tasks and projects updates orthogonally to it, which leaves that layer's outputs on those inputs unchanged. COP builds on the activation-space projection of GPM and applies it to the subspaces of past forget sets, so that the quantity it protects is the value each earlier unlearning objective reached.
 
\paragraph{Gradient projection in unlearning.}
Several unlearning methods project the forget update to protect retained data within a single request. PGU~\citep{hoang2024learn} computes a GPM-style basis of each layer's input activations on the retained data and projects the forget gradient onto its orthogonal complement, and UNSC~\citep{chen2024machine} confines unlearning to the null space of the remaining classes' layer inputs; both target image classifiers.
In LLMs, GU~\citep{zhou2025geometricdisentangelment} projects the update of an arbitrary gradient-based unlearning objective orthogonally to the span of retain gradients, which keeps the retain loss unchanged to first order. These methods protect the retain set but not earlier deletions.
Two methods apply projection across sequential requests. FG-OrIU~\citep{feng2025fgoriu} decomposes layer features into forgetting and remaining subspaces, projects gradients with respect to both, and adds a loss that makes forget features orthogonal to them, expanding the forgetting subspace as requests arrive; it is evaluated on vision models and couples the projection with its own objective.
\citet{lee2026continual} study continual unlearning in text-to-image diffusion models and propose a gradient projection that keeps parameter drift orthogonal to the subspace of concepts semantically related to each target. COP differs from these in targeting LLMs, in protecting the earlier forget sets themselves through cached activation subspaces, and in leaving the unlearning objective unchanged.

\section{Evaluation metrics} \label{app:metrics}
 
Both benchmarks consist of short questions with ground-truth answers, so we evaluate them with a common set of metrics. Forget ROUGE (FR) is the ROUGE-L recall of the ground-truth answer against the model's greedy generation. Forget probability (FP) is the length-normalized probability of the ground-truth answer. Exact memorization (EM)~\citep{tirumala2022memorization} is the fraction of answer tokens that match the model's top prediction given the preceding ground-truth tokens. Extraction strength (ES)~\citep{carlini2021extracting} is one minus the fraction of the answer that must be given as a prefix before the model's greedy predictions reproduce the remainder exactly. After request $k$, forget metrics are averaged over the pooled examples of $\mathcal{F}_1,\ldots,\mathcal{F}_k$. Model utility (MU) is a harmonic mean of answer probability and ROUGE on data the model should retain, with a composition specific to each benchmark and given in its setup, so utilities are not comparable across benchmarks. Forget quality~\citep{maini2024tofu} requires a model trained without the forget data, which is available only for TOFU, and we use it solely for hyperparameter selection (Appendix~\ref{app:tofu}).

\section{Implementation details} \label{app:impl}

\paragraph{Training.}
Every method trains the full model in bfloat16 and processes the requests in order on the same model, with a new optimizer and schedule at every request. At request $k$, each training example pairs one example of $\mathcal{F}_k$ with one drawn uniformly at random from $\mathcal{R}$, so one epoch is one pass over $\mathcal{F}_k$. Following the reproduction setup of \citet{dorna2025openunlearning}, GA, NPO, SimNPO, IDK, and their COP variants train for 10 epochs per request, about 20 optimizer steps on TOFU, using AdamW with weight decay $0.01$, a linear schedule with 10\% warmup, and an effective batch size of 32 (8 per device with 4 accumulation steps). Every objective includes a retain negative log-likelihood term with weight $\lambda=1$, so GA corresponds to gradient difference, and the retain term always uses the original retain answers. Every COP run has an unaugmented counterpart with identical seed, learning rate, loss, data, and training budget, so the two differ only in the projection.

Experiments are run on a single NVIDIA RTX PRO 6000 Blackwell (96 GB) workstation GPU, with an AMD Ryzen Threadripper 7980X CPU and 246 GB of RAM; the full set of runs reported in this paper takes about 300 GPU hours.

\paragraph{Hyperparameters.}
Following the ranges reported by \citet{dorna2025openunlearning}, we search learning rates in $\{1,2,5\}\times10^{-5}$ for all objectives, $\beta\in\{0.05,0.1,0.5\}$ for NPO and IDK, and $\beta\in\{2.5,3.5,4.5\}$ for SimNPO, with forget weight $\gamma=0.125$ and margin $\delta=0$ for SimNPO and $\gamma=1$ otherwise. Each COP variant repeats every grid point of its base objective with the rank cap $r_{\max}\in\{16,64,256\}$, the maximum number of new directions per layer per request. Results in the main tables use $r_{\max}=256$, and the remaining ranks are used for the ablation in Figure~\ref{fig:rank}. For O$^3$, we search learning rates in $\{1,2,3\}\times10^{-4}$ and orthogonality loss weights in $\{0.05,0.1,0.2\}$ with 10 epochs per request. ALKN sweeps the learning rate over $\{1, 3, 10\}\times10^{-5}$, the number of optimizer steps per request over $\{10, 30\}$, and the factor applied to the learning rate after each request over $\{0.5, 1\}$. Its soft-label scale is fixed at 10, its negation scale at $0.8$, the ratio of its low to high learning rate at $0.1$, and the fraction of parameters left unmasked at the last step at $0.3$, with the same batch, weight decay, and warmup as the other objectives. Every grid point runs once, with seed 0. The energy threshold $\tau$ and the selection criterion differ between benchmarks and are given below.

\subsection{TOFU}
\label{app:tofu}

\paragraph{Benchmark.}
TOFU~\citep{maini2024tofu} consists of 4{,}000 synthetic question-answer pairs, 20 for each of 200 fictitious authors whose biographies were generated by GPT-4, so none of this knowledge can appear in pretraining data. Because the authors are fictitious, the benchmark also provides a reference model fine-tuned only on the retain set, which approximates the ideal outcome of unlearning; we use the \texttt{retain95} reference models released by \citet{dorna2025openunlearning}. Questions about real authors and world facts, used in MU, probe whether general knowledge survives unlearning.

\paragraph{Hyperparameter selection.}
COP additionally searches $\tau\in\{0.90,0.95,0.99\}$. For each method and variant, we select the configuration with the highest Continual Retain-Forget Score (CRFS), the harmonic mean between model utility and forget quality, both measured after the final request.
Forget quality (FQ)~\citep{maini2024tofu} is the $p$-value of a two-sample Kolmogorov-Smirnov test between the truth ratio distributions of the unlearned model and the retain-only reference model on the forget set. The truth ratio compares the likelihood the model assigns to paraphrased correct answers against perturbed incorrect ones, so a high FQ means the two distributions are statistically indistinguishable, while a low FQ indicates that the unlearned model still behaves differently from a model that never saw the forgotten authors. We compute FQ over the union of all forget requests, so it penalizes configurations that let earlier deletions resurface. The harmonic mean is dominated by the lower of its two terms, so CRFS discards configurations that either collapse utility or barely forget, which a single metric would favor. Since FQ is a $p$-value, it serves well as a selection criterion but is not directly comparable as a measure of forgetting strength, so we report memorization metrics in the main text instead.

\subsection{RWKU} \label{app:rwku}

\paragraph{Data.}
We load the \texttt{forget\_level2} configuration of RWKU, whose single split contains 2,879 question-answer probes, each annotated with the person it concerns. The 20 forget people are sampled from the distinct people sorted by name, independently of the training seed, so they are the same for every method and run, and consecutive groups of four form $\mathcal{F}_1,\ldots,\mathcal{F}_5$. We use every probe of a selected person, without filtering probes by whether the base model answers them correctly before unlearning.

\paragraph{Prompt and generation.}
Each probe is posed with the RWKU question prompt as a single user turn, and the assistant turn begins with a prefilled \texttt{Answer:} that the model completes with the answer. The prefix is excluded from every loss and metric. Generation is greedy and stops at the first newline or after 32 new tokens, and ROUGE is computed with stemming. Cumulative forget metrics average per probe over the pooled requests.

\paragraph{Method-specific settings.}
IDK replaces each forget answer with a refusal drawn from a fixed list, assigned deterministically by example index so that a probe receives the same refusal whenever it is seen, and prefixed with \texttt{Answer:}. O$^3$ trains LoRA adapters of rank 8, scaling factor 32, and dropout $0.05$ on the attention and MLP projections, with a per-device batch of 1, 4 accumulation steps, and weight decay $0.1$. It trains on the same refusal targets, uses no retain data, and configures its out-of-distribution detector with the TOFU settings, scored on the raw question text.

\paragraph{Hyperparameter selection.}
COP fixes $\tau=0.99$, giving three COP runs per baseline configuration. In total the grid has 30 baseline runs, 90 COP runs, 9 O$^3$ runs, 12 ALKN runs, and one evaluation of the unmodified base model at every request. Forget quality requires a reference model trained without the forget data, which RWKU cannot provide because its knowledge comes from pretraining. For each method and variant, we therefore select the configuration with the highest harmonic mean of model utility and one minus exact memorization, measured after the final request, with exact memorization computed over the union of all requests.

\section{Effect of subspace rank cap}
\label{app:rank}

The rank cap $r_{\max}$ bounds how many directions each request adds to the protected subspace. A larger cap protects earlier deletions and retained behavior more completely, but it also restricts the directions available to later requests, which could make new deletions harder to achieve. We study this trade-off by varying $r_{\max}$ on TOFU with Llama-3.2-1B-Instruct, keeping the stream of five requests and the remaining setup of Section~\ref{sec:tofu}. 

\begin{figure}[h]
    \centering
    \includegraphics[width=\columnwidth]{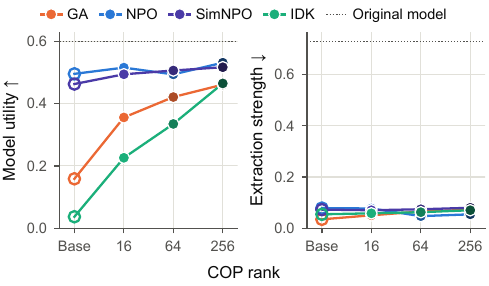}
    \caption{\textbf{Larger protected subspaces improve utility without weakening forgetting.} Final model utility (left) and extraction strength (right) on TOFU with Llama-3.2-1B-Instruct, as a function of the COP rank cap $r_{\max}$. ``Base'' denotes each objective without COP, and dotted lines mark the original model. Utility grows with rank for all methods, sharply for GA and IDK and moderately for NPO and SimNPO, while extraction strength remains far below original model at every rank.}
    \label{fig:rank}
\end{figure}

Figure~\ref{fig:rank} shows that final utility grows with $r_{\max}$ for every objective, sharply for GA and IDK and moderately for NPO and SimNPO, as protecting more directions further limits interference with earlier deletions and retained knowledge. Extraction strength, in contrast, stays nearly constant across ranks and far below that of the original model, so the restriction on later updates does not prevent new deletions from taking effect. Within the range we test, increasing the rank cap thus improves utility at no cost to forgetting, and its main cost is the memory needed to store the protected basis.

\section{Proof of Theorem~\ref{thm:exact}}
\label{app:proof-exact}
 
Fix $k\ge i$ and a layer with weight $W$. Every update $\Delta W$ applied during a step $n>i$ satisfies $\Delta W\,V_{n-1}=0$, and $\mathcal{S}_i\subseteq\mathcal{S}_{n-1}$, so $\Delta W\,x=0$ for every $x\in\mathcal{S}_i$. Summing these updates gives, for all $x\in\mathcal{S}_i$,
\begin{equation}
  \label{eq:proof-accumulated}
  W(\theta_k)\,x=W(\theta_i)\,x
\end{equation}
independently of the step sizes.
 
The forward pass on $z$ is a finite computation in which each operation either applies a trainable linear weight to its input or is a fixed function of its inputs, such as attention, normalization, or a nonlinearity, since every trainable parameter is a linear weight. We show by induction over this computation, in the order in which it is evaluated, that every intermediate value is the same under $\theta_k$ and $\theta_i$.

The input tokens are identical. A fixed operation receives identical inputs and therefore returns an identical value. A linear operation receives, by the induction hypothesis, the same input $x$ under both parameters, and $x\in\mathcal{S}_i$ by assumption, so its output is unchanged by \eqref{eq:proof-accumulated}. The model output is the last of these values, hence $f_{\theta_k}(z)=f_{\theta_i}(z)$. \hfill$\qed$

\end{document}